\documentclass[11pt]{article}

\usepackage[final]{acl}

\usepackage{times}
\usepackage{latexsym}

\usepackage[T1]{fontenc}

\usepackage[utf8]{inputenc}

\usepackage{microtype}

\usepackage{inconsolata}
\usepackage{graphicx}
\usepackage{amsmath}
\usepackage{amssymb}
\usepackage{xcolor}
\usepackage{hyperref}
\title{Explainable Identification of Hate Speech towards Islam using Graph Neural Networks}

\author{
 \textbf{Azmine Toushik Wasi}
\\
Shahjalal University of Science and Technology, Bangladesh
\\
\texttt{azmine32@student.sust.edu}
}

\begin{document}
\maketitle
\begin{abstract}
Islamophobic language on online platforms fosters intolerance, making detection and elimination crucial for promoting harmony. Traditional hate speech detection models rely on NLP techniques like tokenization, part-of-speech tagging, and encoder-decoder models. However, Graph Neural Networks (GNNs), with their ability to utilize relationships between data points, offer more effective detection and greater explainability. In this work, we represent speeches as nodes and connect them with edges based on their context and similarity to develop the graph. This study introduces a novel paradigm using GNNs to identify and explain hate speech towards Islam. Our model leverages GNNs to understand the context and patterns of hate speech by connecting texts via pretrained NLP-generated word embeddings, achieving state-of-the-art performance and enhancing detection accuracy while providing valuable explanations. This highlights the potential of GNNs in combating online hate speech and fostering a safer, more inclusive online environment.

\textcolor{red}{\textit{\textbf{Disclaimer:} This manuscript may contain examples of hateful or offensive language, as it discusses hate speech in the context of detection and analysis. These instances are included strictly for research purposes and do not reflect the authors' views. Reader discretion is advised.}}
\end{abstract}


\section{Introduction} \label{sec:intro}
Detecting and eliminating hate speech on social media platforms is of utmost importance for the promotion of harmony and tranquility in society \cite{Rawat202efwws4, Kovcs2021fwefew, Davidson2017AutomatedHS}. The escalating presence of hate speech specifically targeting Islam or Muslim communities on online discussion platforms is a growing concern \cite{2022MakingMT}. This form of hate speech not only fosters an environment of intolerance and hostility but can also have severe psychological impacts on individuals and communities, leading to real-world violence and discrimination \cite{Saha20194ge4tg}.

\begin{figure}[t]
\centering {\includegraphics[scale=0.3]{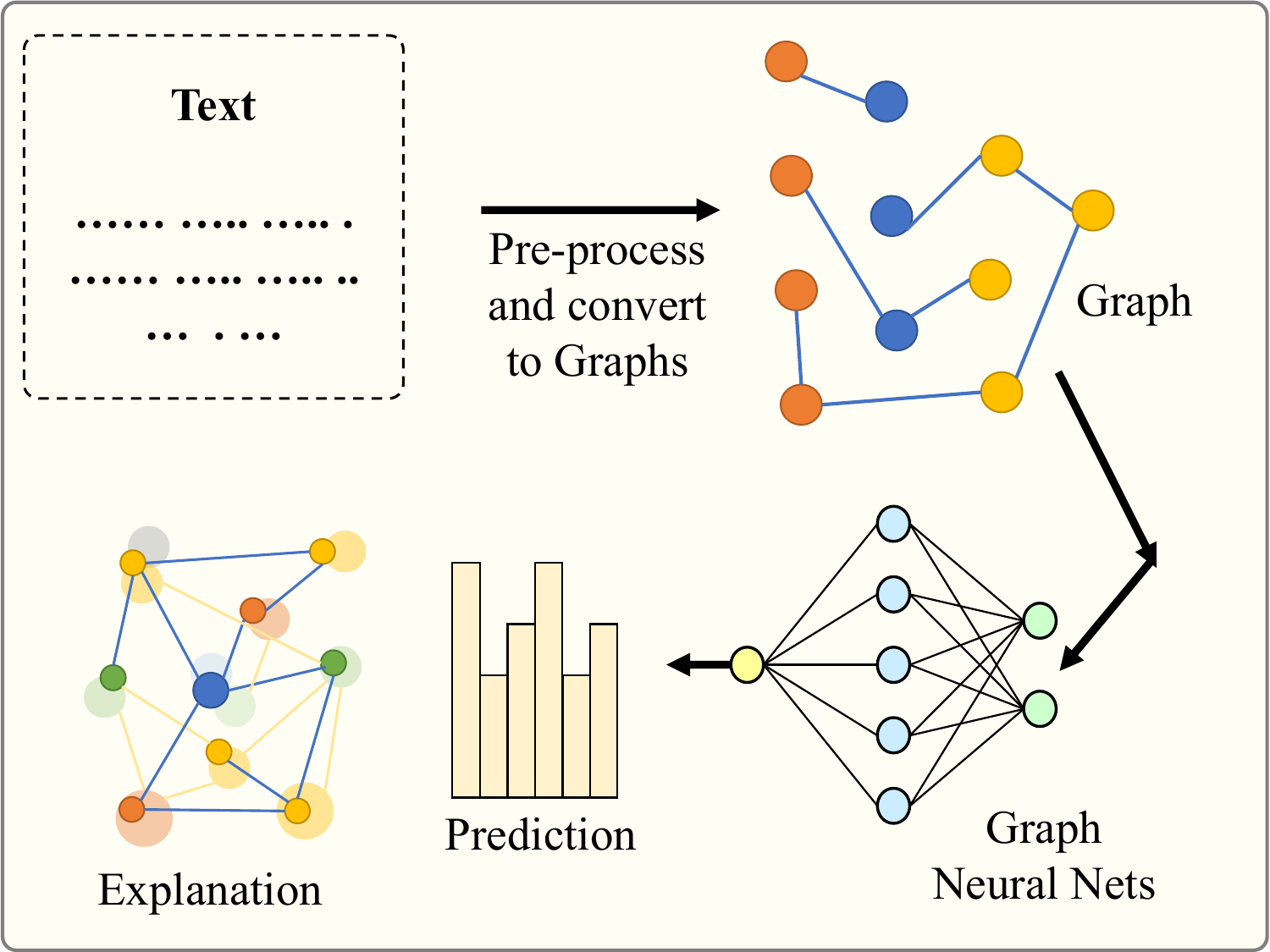}} \label{fig:what}
\caption{Our approach of Hate Speech towards Islam using GNNs}
\vspace{-4mm}
\end{figure}

To address this issue, researchers have increasingly turned to advanced technologies; using text-processing approaches in AI. Natural Language Processing (NLP) techniques are frequently employed for hate speech detection, with some offering severity assessment of hate speech \cite{Badjatiya2017DeepLFm, Mozafari2020HateSD,Schmidt2017ASO, Cao2020DeepHateHS}. These methods utilize sophisticated algorithms to analyse vast amounts of textual data, identifying patterns and features indicative of hate speech. For instance, deep learning models, like recurrent neural networks (RNNs), can learn complex representations of text data, enabling them to detect subtle and context-dependent instances of hate speech \cite{Schmidt2017ASO}. 
Modern NLP techniques, on the other hand, can enhance these models by providing richer linguistic insights. Tokenization, part-of-speech tagging, and named entity recognition are just a few NLP techniques that help in breaking down and understanding the text's structure and meaning. Moreover, the integration of latest NLP model and transformers, like BERT \cite{gdszgzsgbsrgbrs} and GPT \cite{radford2019language,brown2020languagemodelsfewshotlearners}, has significantly improved the ability of models to understand context \cite{ferrando-etal-2023-explaining}, sarcasm \cite{A20213wt43watgawt}, and implicit hate speech \cite{Cao2020DeepHateHS,Mozafari2020HateSD}, which are often challenging to detect. Another interesting approach is to use human-centric perspectives of AI using some benchmark dataset \cite{wasi2024exploringbengalireligiousdialect,wasi2024diaframewesefew}.

Researchers have tried to employ GNNs in hate speech classification \cite{drhrsdhsdh235253,hebert2023predicting, Blc2021HateSA}, but still needs more focus on this area. Despite their potential, GNNs have not been actively employed for the purpose of interpretable identification of hate speech, particularly in Islamic contexts. Islamophobic\footnote{In this work, the terms "hate speech towards Islam" and "Islamophobic hate speech" is used interchagably.} content often exhibits close word choices and hate speakers from the same community, which GNNs can leverage to reveal and explain patterns, alongside impressive classification scores.

\begin{figure*}[t] 
\centering {\includegraphics[scale=0.575]{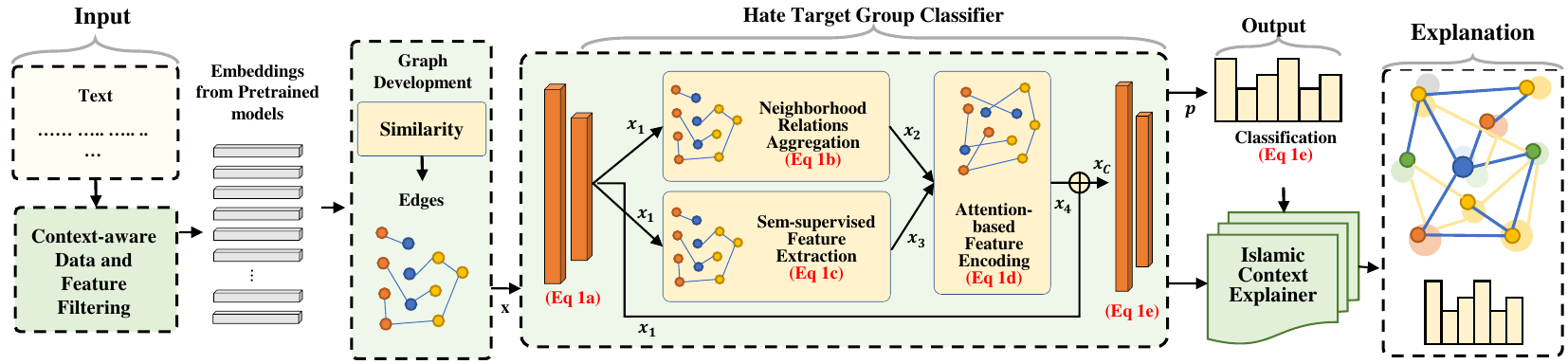}}
\caption{Our framework for Explainable Identification of Hate Speech towards Islam using GNNs.}\label{fig:main}
\vspace{-4mm}
\end{figure*}

In this study, we introduce a novel approach employing graph neural networks for the identification and explication of hate speech directed at Islam (\texttt{XG-HSI}), as demonstrated in Figure \ref{fig:main}. We pre-process the dataset to focus on Islamic contexts, utilize pretrained NLP models for word embeddings, establish connections between texts, and employ a series of graph encoders for hate speech target identification, which achieves state-of-the-art performance. 

\section{Background}
Graph Neural Networks (GNNs) are powerful neural networks designed for processing non-Euclidean\footnote{In this context, non-Euclidean data refers to data that lacks a regular grid structure, like graphs where nodes (data points) are connected by edges (relationships). Unlike Euclidean data such as images or sequences, which follow a structured grid or linear pattern, non-Euclidean data is irregular and interconnected, such as a social network where users (nodes) have varying numbers of connections (edges). GNNs are designed to process this complex graph structure, capturing relationships and patterns that traditional neural networks cannot handle effectively \cite{scardapane2024alicesadventuresdifferentiablewonderland}.} data organized in complex, interconnected graphs \cite{scardapane2024alicesadventuresdifferentiablewonderland,waikhom2021graph}. Using their ability to utilize relations between different data points \cite{Xu2018HowPA}, GNNs have shown tremendous promise in text classification and detection tasks \cite{Lu2020VGCNBERTAB,Zhang2020EveryDO,Pal2020MultiLabelTC}. GNNs have the ability to enhance hate speech detection on social media by modeling complex relationships between users and content, capturing contextual information from interactions. They propagate information across the network, identifying coordinated and evolving hate speech patterns. We also present a case study in Section \ref{sec:case-study} to illustrate how incorporating related information enhances the process. Recent graph-based approaches uses LLM to develop knowledge graphs BanglaAutoKG \cite{wasi-etal-2024-banglaautokg}, World Knowledge in Implicit Hate Speech Detection \cite{lin-2022-leveraging}, and HRGraph \cite{wasi-2024-hrgraph}.

In this work, we adopted a general bag of words-based approach to create graphs, without LLMs. By integrating with pretrained NLP models, GNNs leverage contextual word embeddings to better understand the subtleties of hate speech. This combined approach improves the accuracy, context-awareness, and adaptability of detection systems, making them more effective in identifying hate speech directed at Islam and potentially generalizing to other targeted groups.

\section{Methodology}
\subsection{Notations}
Let a graph $\mathcal{G}=(\mathcal{V}, \mathcal{E}, X)$, where $\mathcal{V}$ represents nodes, $\mathcal{E}$ denotes edges. We also define $N$ and $M$ as the numbers of nodes and edges, respectively. Each node $v$ is associated with a feature $x_i \in \mathbb{R}^F$, and the node feature matrix for the entire graph is denoted as $X \in \mathbb{R}^{N \times F}$, where $F$ represents the feature vector length. In our approach, each content\footnote{Each content denotes the full post, that was labelled as hate speech or not.} denotes a node, contextual similarity between two nodes is denoted by an edge and word embeddings are node features of the graph. The task involves a node classification task to detect hate speech and Islamophobic content.

\subsection{Data Pre-Processing}
Initially, the dataset was filtered to focus on hate speech targeting Islam. Next, pretrained NLP models is applied to the text to obtain word embeddings $X$ as node features for all nodes $\mathcal{V}$. Edges $\mathcal{E}$ are determined using cosine similarity between embeddings with a threshold of 0.725. Subsequently, GNN is applied for the classification task.

\subsection{Graph Encoder}
After data pre-processing, every data point\footnote{the BERT-embeddings, denoting the features of each sentence.} $x \subset X$ undergoes a series of transformations to get output $p$. First, it is processed by a linear layer producing $x_1$ (Equation \ref{eq:1}). 

\begin{equation}
    x_1 = Wx+b  \label{eq:1}
\end{equation}

Subsequently, $x_1$ is passed into two initial graph encoders to aggregate neighborhood information, feature extraction, and yield $x_2, x_3$ utilizing $\mathcal{G}$ 
and concatenated to $x_{23}$ (Equation \ref{eq:2},\ref{eq:3}, \ref{eq:3.1}). 
Here in Equation \ref{eq:2}, we aggregate features from a node's local neighborhood, to learn different characteristics \cite{hamilton2018inductive}. 
In Equation \ref{eq:3} and \ref{eq:3.1}, we use a semi-supervised learning on graph-structured data, employing an efficient variant of convolutional neural networks that operate directly on graphs \cite{kipf2017semisupervised}.

\begin{equation}
{x}_2  ={W}_1 {x}_1+ {W}_2 \cdot \operatorname{mean}_{j \in \mathcal{N}(i)} {x}_1 \label{eq:2}
\end{equation}
\begin{equation}
 x_3 = W_1 x_{1_i} + W_2 \sum x_{1_j} \label{eq:3}
\end{equation}
\begin{equation}
 x_{23}  = concat(x_2,x_3);   \label{eq:3.1}
\end{equation}

Here, $\mathcal{N}$ is the set of neighbouring nodes.
Following this, $x_{23}$ is passed through another graph layer employing attention-based feature extraction,
 utilizing masked self-attentional layers to implicitly assign different weights to nodes in a neighbourhood \cite{veličković2018graph}, producing $x_4$ (Equation \ref{eq:4} and \ref{eq:4.1}).

\begin{equation}
        {x}_4 =\alpha_{i, i} \Theta {x}_{23_i}+\sum \alpha_{i, j} \Theta {x}_{23_j}  \label{eq:4}
\end{equation}

\begin{equation}
\small
\alpha  =\frac{\exp \left(\operatorname{LeakyReLU}\left({a}^{\top}\left[{\Theta} {x} \| {\Theta} {x}_{23_j}\right]\right)\right)}{\sum \exp \left(\operatorname{LeakyReLU}\left({a}^{\top}\left[{\Theta} {x}_{23_i} \| {\Theta} {x}_{23_k}\right]\right)\right)} \label{eq:4.1}
\end{equation}

Here, $\theta$ refers to trainable model weights. $\alpha$ is the attention value, calculated by the equation mentioned.

Finally, $x_4$ is passed through a final linear layer to obtain logits $p_l$, which are then subjected to a softmax operation to derive probabilities $p$ (Equation \ref{eq:5} amd \ref{eq:5.1}).

\begin{equation}
x_{c} = concat(x_1,x_4); p_l = Wx_c+b   \label{eq:5}
\end{equation}

\begin{equation}
   p = softmax(p_l)  \label{eq:5.1}
\end{equation}

An illustration of the network in presented in Figure \ref{fig:main}.

\subsection{Loss Function} 
Cross Entropy loss \cite{mao2023crossentropylossfunctionstheoretical} is designed to minimize the difference between the predicted probabilities and true values, as follows: 
\begin{equation}
\small
l_{CE}=-\frac{1}{n} \sum_{i=1}^n\left(p_i^{\prime} \log \sigma\left(p_i\right)+\left(1-p_i^{\prime}\right) \log \left(1-\sigma\left(p_i\right)\right)\right)
\end{equation}

\subsection{Graph Explanation} 
GNNExplainer \cite{ying2019gnnexplainer} is used to derive explanations from the graph encoder network for interpreting the results and find underlying relations and causation. It works by taking a trained GNN model and its predictions as input, and returns explanations in the form of compact subgraph structures and subsets of influential node features. This model-agnostic approach can explain predictions of any GNN-based model on various graph-based machine learning tasks, including node classification, link prediction, and graph classification. GNNExplainer formulates explanations as rich subgraphs of the input graph, maximizing mutual information with the GNN's predictions. It achieves this by employing a mean field variational approximation to learn real-valued graph masks that select important subgraphs and feature masks that highlight crucial node features. Through this process, GNNExplainer offers insights into the underlying reasoning of GNN predictions, enhancing model interpretability and facilitating error analysis.

\section{Experiments}
\subsection{Experimental Setup}
\textbf{Dataset.}
We use HateXplain \cite{Mathew2020HateXplainAB}, a benchmark hate speech dataset designed for addressing bias and interpretability. The dataset has hate speech targets labelled. We use this labelling to collect only Muslim-focused sentences and created a subset to work on this project. We have used a 6:2:2 train, validation and test split in our work.

\noindent
\textbf{Baselines.}
The baseline models are: \texttt{CNN-GRU}, \texttt{BiRNN} \cite{bi-rnn}, \texttt{BiRNN-HateXplain} \cite{Mathew2020HateXplainAB}, \texttt{BERT} \cite{gdszgzsgbsrgbrs}, \texttt{BERT-HateXplain} \cite{Mathew2020HateXplainAB}. Mentioned HateXplain-based models are fine-tuned on HateXplain dataset \cite{Mathew2020HateXplainAB}.

\noindent
\textbf{Implementation Details.}
Hugging Face transformers library \cite{wolf-etal-2020-transformers} is used to get embeddings from pre-trained \texttt{BERT} (\texttt{bert-base-uncased}) \cite{gdszgzsgbsrgbrs} and \texttt{BiRNN} \cite{bi-rnn}. The model is trained for 200 epochs with a learning rate of 0.001, using Adam optimizer. The experimental results in Table 1 show that our model achieves remarkable performance comparing to benchmarks with explaining occurring phenomenons.We utilized a single layer for each type of GNN, with a maximum tokenization length of 512 in the tokenizer and length of BERT embeddings ($F$) set to 128.

\begin{table}[t]
\centering
\caption{Experimental Results ($\uparrow$)}\label{table:res}
  \begin{tabular}{lcc} 
  \hline
  \small
  Model & Accuracy & Macro F1 \\ 
   \hline
   CNN-GRU  & 0.628 & 0.604 \\
   BiRNN  & 0.591 & 0.578 \\ 
   BiRNN-HateXplain  & 0.612 & 0.621 \\ 
   BERT  & 0.692 & 0.671 \\ 
   BERT-HateXplain & 0.693 & 0.681\\ \hline
   \textbf{\texttt{XG-HSI-BiRNN} (Ours)} &  0.742 & 0.737 \\
   \textbf{\texttt{XG-HSI-BERT} (Ours)} &  \textbf{0.751} & \textbf{0.747} \\\hline
  \end{tabular}
  \vspace{-4mm}
\end{table}

\subsection{Experimental Results}
Table \ref{table:res} shows the performance of various models in detecting hate speech, highlighting accuracy and Macro F1 metrics. Traditional models like \texttt{CNN-GRU} and \texttt{BiRNN} show lower performance, with \texttt{BiRNN-HateXplain} offering slight improvements. BERT-based models perform better, particularly \texttt{BERT-HateXplain}. However, our proposed models, \texttt{XG-HSI-BiRNN} and \texttt{XG-HSI-BERT}, significantly outperform all others, with \texttt{XG-HSI-BERT} achieving the highest accuracy (0.741) and Macro F1 (0.747). These results demonstrate the superior effectiveness of our dual GNN approach in hate speech detection.

\section{Graph Explanation Case Study} \label{sec:case-study}
For a given post, \textbf{\textit{"How is all that awesome Muslim diversity going for you native germans? You have allowed this yourselves. If you do not stand and fight against this. You get what you asked for what you deserve!"}}, the predicted classification was offensive towards Islam. As per the explainer (Figure \ref{fig:case-sty}), the neighbouring and self-tokens\footnote{Each sentence was tokenized, and then we collected embeddings from BERT as features. Those numbers in the figure denotes to particular token, used in tokenization.} helped to classify this as offensive to Muslims are \textbf{\textit{fight, Muslim diversity, brooks, \#\#rish, donald, syrian, schultz, typed}}.  The text's association of \textit{"Muslim diversity"} with potential blame and its confrontational tone in phrases like \textit{"stand and fight against this,"} combined with neighbouring tokens like syrians, brooks, syrians denoted negative sentiment. More detailed analysis is added in Appendix \ref{sec:ex-case-study}.

\begin{figure}[t]
\centering {
\includegraphics[scale=0.475]{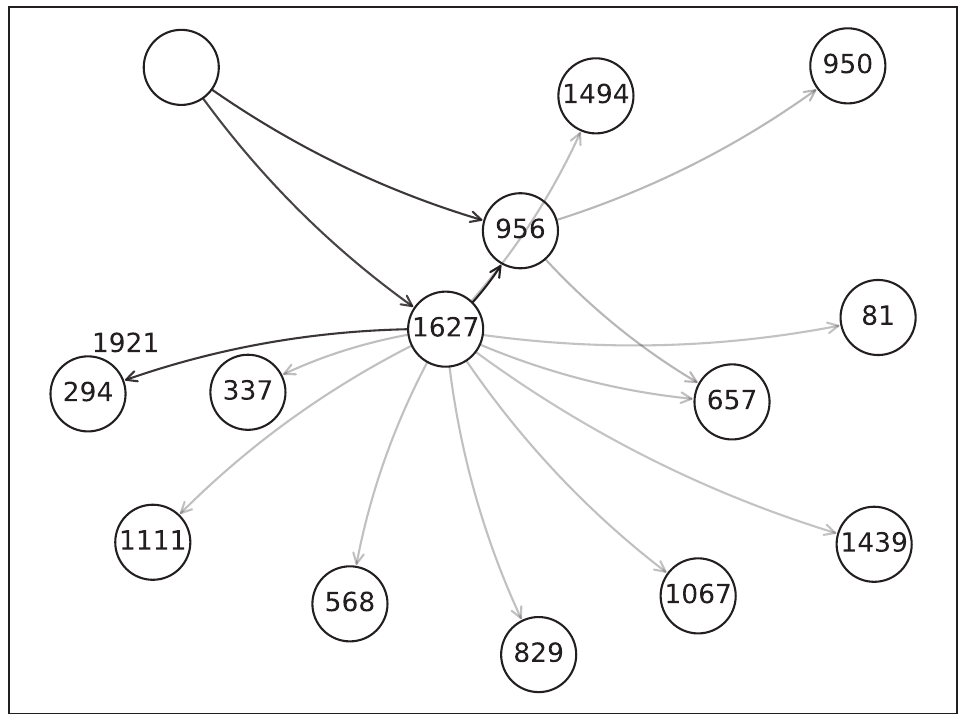}
} 
\caption{Explanation Graph} \label{fig:case-sty}
\vspace{-4mm}
\end{figure}

\section{Discussion}
We believe that our study not only addresses the immediate challenge of identifying and explaining hate speech directed at Islam but also recognizes the broader impact of hate speech propagation on online platforms. The proliferation of Islamophobic language fosters intolerance, division, and hostility within communities, perpetuating harmful stereotypes and prejudices. By leveraging GNNs in our \texttt{XG-HSI} framework, we not only detect hate speech but also provide explanations for its occurrence, shedding light on the underlying factors driving such behaviour.
GNNs excel in capturing complex relationships and patterns within data, enabling them to effectively identify instances of hate speech and elucidate the contextual nuances surrounding them. By leveraging the inherent structure of social networks and textual data, our approach offers a comprehensive understanding of how hate speech propagates in online discourse.

In future research, exploring the integration of multimodal data sources, such as images and videos, could enhance the robustness of hate speech detection models, particularly in detecting nuanced forms of Islamophobic content. Additionally, investigating the dynamic nature of online communities and incorporating temporal aspects into GNN architectures could provide deeper insights into the evolution of hate speech propagation and enable more proactive interventions to counter its spread.

\section{Conclusion}
Identifying and addressing Islamophobic hatred on social media is crucial for achieving harmony and peace. This research presents a novel method using GNNs to detect hate speech towards Islam. Empirical findings demonstrate that our model achieves exceptional performance, significantly outperforming all others, with \texttt{XG-HSI-BERT} achieving the highest accuracy (0.741) and Macro F1 (0.747). Explainability aspect of this approach is also very promising, as it provides insights into both correlations and causation. This further highlights the potential of GNNs in combating online hate speech and fostering a safer, more inclusive online environment.

\section*{Limitations}
The limitations of our study include the use of only one dataset, which, while sufficient for this initial exploration, should be expanded upon in future research to validate and extend our findings. Additionally, while Graph Neural Networks (GNNs) are known to be computationally intensive, especially with large-scale datasets, the relatively limited number of hate speech keywords suggests that GNNs may still be highly effective. Furthermore, more efficient GNN training methods are now available such as G3 \cite{10.1145/3589288} and Graphite \cite{10.1145/3470496.3527403}, which address some of the computational challenges in future applications.

\section*{Ethical Implications}
Our work on using GNNs to detect hate speech targeting Islam carries significant ethical responsibilities. We focus on minimizing biases in the model to ensure fair treatment of all groups, emphasizing the need for transparency in how the model arrives at its decisions. By using interpretable GNN methods, we strive to provide clear explanations for the model's classifications, allowing for greater accountability. We also acknowledge the potential risks of misuse and take steps to prevent these, adhering to ethical guidelines that respect privacy and avoid unjust censorship.

\section*{Societal Implications}
The societal impact of our work lies in its potential to create a safer online environment by effectively identifying and mitigating Islamophobic content. By enhancing the detection accuracy and providing clear explanations for the identified hate speech, our model contributes to fostering more inclusive and respectful online communities. Additionally, our work highlights the importance of combating digital hate speech, which can lead to real-world harm. We aim to empower platforms and policymakers with tools that uphold freedom of expression while curbing harmful rhetoric, thus promoting social harmony and understanding.

\section*{Potential Risks}
The application of our model presents several risks. One major concern is the potential for model misclassification, which could lead to false positives or negatives, impacting users unfairly. Additionally, there is a risk of over-reliance on automated systems, which might not capture nuanced contexts and could inadvertently suppress legitimate speech. Annotation errors can also induce bias \cite{sap-etal-2019-risk}, but as we used a previously peer-reviewed benchmark dataset, we hope those type of concerns are already addressed.

\section*{Acknowledgements}
I express my sincere gratitude to the Computational Intelligence and Operations Laboratory (CIOL) for all their support. This work was presented at the Muslims in ML workshop (non-archival) at NeurIPS 2023, and I thank them for their reviews, support, and the opportunity to present. I also extend my appreciation to all the reviewers for their valuable suggestions to improve the work.


\bibliography{custom}

\clearpage
\newpage

\appendix
\section{Extended Explanation of Case Study} \label{sec:ex-case-study}

The task of detecting and classifying offensive content, especially hate speech, is inherently complex due to the nuanced and often implicit nature of such language. In the example provided—"How is all that awesome Muslim diversity going for you native Germans? You have allowed this yourselves. If you do not stand and fight against this. You get what you asked for, what you deserve!"—the model identified the post as offensive toward Islam. This classification was aided by analysing specific tokens and their relationships within the text using a Graph Neural Network (GNN) framework, particularly with the GNNExplainer \cite{ying2019gnnexplainer} method.

As discussed above in Section \ref{sec:intro}, GNNs excel in tasks where the relationships between data points are as critical as the data points themselves. In the context of hate speech detection, GNNs can capture the intricate web of semantic and syntactic connections between words, phrases, and even larger text segments. This capability allows the model to consider not just isolated words but also the context in which they appear, making it particularly powerful for understanding language that may be implicitly biased or offensive.

In this example presented in Figure \ref{fig:case-sty}, the GNNExplainer was employed to determine which tokens—both in isolation and in combination with their neighbouring tokens—contributed to the model’s decision to classify the post as offensive. The key tokens identified, such as "fight," "Muslim diversity," and "Syrian," are not inherently negative but, when analysed in context, reveal an underlying sentiment of hostility and blame. The phrase "stand and fight against this" suggests a confrontational stance, while the juxtaposition of "Muslim diversity" with a directive to "stand and fight" subtly frames the diversity as a threat. The mention of "Syrian" further adds to the narrative by invoking a specific group, which, in the context of the surrounding words, contributes to a negative sentiment.

GNN-based explainers are particularly effective because they allow us to visualize and interpret the model's decision-making process by highlighting the most influential tokens and their connections. This interpretability is crucial in sensitive applications like hate speech detection, where understanding why a model made a certain decision can help in refining the model, addressing potential biases, and ensuring that it aligns with ethical guidelines. 
Moreover, by using a GNN-based approach, the model can weigh the significance of different parts of the text more effectively than traditional linear models. The graph structure allows the model to account for the interactions between words and their broader context, providing a more holistic understanding of the text. This is particularly important in hate speech detection, where context often determines whether a statement is offensive.

\begin{figure}[t]
\centering {
\includegraphics[scale=0.475]{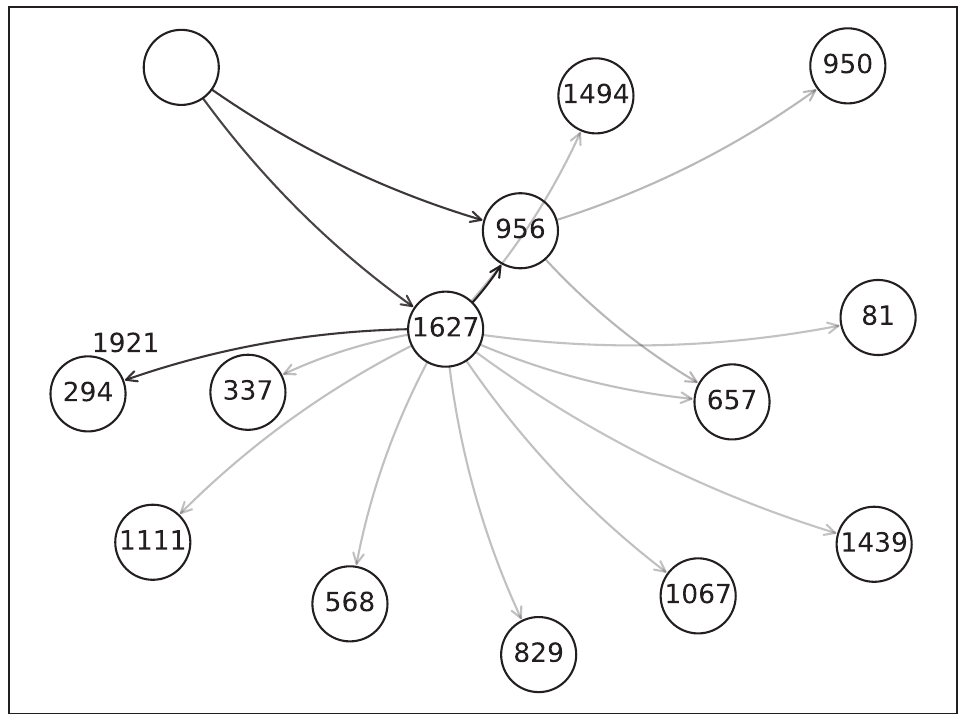}
}
\caption{Explanation Graph} \label{fig:case-sty}
\end{figure}

\end{document}